\documentclass[a4paper,twoside]{article}

\usepackage[utf8]{inputenc}     
\usepackage[T1]{fontenc}        
\usepackage{hyperref} 
\usepackage{graphicx} 
\usepackage{amsfonts}           
\usepackage{cite}               
\usepackage{amsmath,amssymb,bm} 
\usepackage{mathtools}          
\usepackage{siunitx}            
\usepackage{nicefrac}           
\usepackage{booktabs}           

\usepackage{epsfig}
\usepackage{subcaption}
\usepackage{calc}
\usepackage{amstext}
\usepackage{amsthm}
\usepackage{multicol}
\usepackage{pslatex}
\usepackage{apalike}
\usepackage{algorithm2e}
\usepackage[bottom]{footmisc}
\usepackage{SCITEPRESS}     

\begin{document}

\title{Ablation Study to Clarify the Mechanism of Object Segmentation in Multi-Object Representation Learning}

\author{\authorname{Takayuki Komatsu\sup{1}, Yoshiyuki Ohmura\sup{1} and Yasuo Kuniyoshi\sup{1}}
\affiliation{\sup{1}The University of Tokyo}
\email{\{komatsu, ohmura, kuniyosh\}@isi.imi.i.u-tokyo.ac.jp}
}


\abstract{
Multi-object representation learning aims to represent complex real-world visual input using the composition of multiple objects.
Representation learning methods have often used unsupervised learning to segment an input image into individual objects and encode these objects into each latent vector.
However, it is not clear how previous methods have achieved the appropriate segmentation of individual objects.
Additionally, most of the previous methods regularize the latent vectors using a Variational Autoencoder (VAE)\cite{VAE}\cite{Survey_MO_VAE}.
Therefore, it is not clear whether VAE regularization contributes to appropriate object segmentation.
To elucidate the mechanism of object segmentation in multi-object representation learning, we conducted an ablation study on MONet \cite{MONet}, which is a typical method.
MONet represents multiple objects using pairs that consist of an attention mask and the latent vector corresponding to the attention mask.
The attention masks are generated from the input image, are mutually exclusive and cover the entire input image region.
Each latent vector is encoded from the input image and attention mask.
Then, the component image and attention mask are decoded from each latent vector.
The loss function of MONet consists of 1) the sum of reconstruction losses between the input image and decoded component image, 2) the VAE regularization loss of the latent vector, and 3) the reconstruction loss of the attention mask to explicitly encode shape information.
For the first loss, each reconstruction loss is weighted by the attention mask pixel by pixel.
Thus, for each pixel, the smaller the reconstruction loss, the larger the attention mask.
We conducted an ablation study on these three loss functions to investigate the effect on segmentation performance.
Our results showed that the VAE regularization loss did not affect segmentation performance and the others losses did affect it.
Based on this result, we hypothesize that it is important to maximize the attention mask of the image region best represented by a single latent vector corresponding to the attention mask.
We confirmed this hypothesis by evaluating a new loss function with the same mechanism as the hypothesis.
}

\onecolumn \maketitle \normalsize \setcounter{footnote}{0} \vfill

\section{\uppercase{Introduction}}\label{sec:introduction}

The goal of multi-object representation learning is to represent a complex real-world visual scene that contains multiple objects.
Even when there are only a few objects, their combination of them can be very diverse.
Thus, it can be difficult to represent a visual scene that contains multiple objects using a single latent vector \cite{MONet}.
Thus, the basic approach in multi-object representation learning is to decompose the scene into individual objects and then represent each object using a corresponding latent vector.

Previous methods usually adopted the simultaneous unsupervised learning of object segmentation and encoding segmented images into multiple latent vectors via reconstruction.
However, it is not clear how these methods achieved the appropriate segmentation of individual objects.
Additionally, most of these methods used not only a loss function to reconstruct the input image but also the Variational Autoencoder (VAE) \cite{VAE} regularization loss on the latent vectors \cite{Survey_MO_VAE}.
The purpose of adopting the VAE regularization loss is to disentangle each element of each latent vector \cite{Understanding_VAE}.
However, it is not clear whether VAE regularization contributes to appropriate object segmentation.
Understanding the mechanism of object segmentation in multi-object representation learning is an important issue for improving the performance of existing multi-object representation learning methods.
Additionally, it helps to understand the relationship between multi-object representation learning methods and other related methods, such as unsupervised segmentation methods without reconstruction \cite{Invariant_Clustering}.

In a previous study, researchers investigated the mechanism of object segmentation in multi-object representation learning.
Engelcke et al. hypothesized that it is important for appropriate object segmentation to control "reconstruction bottlenecks," which is the capacity of a single latent vector to reconstruct the image.
Then, they investigated the relationship between object segmentation performance and the dimension of each latent vector in GENESIS \cite{GENESIS}, which was a state-of-the-art method for multi-object representation learning \cite{Analysis_Bottleneck}.
They showed that if the reconstruction bottlenecks are too narrow or too wide, the segmentation performance can be degraded.
However, it is still not clear which loss function contributes to appropriate object segmentation.
An ablation study to reduce the used loss functions is important in order to identify the minimum mechanism required for appropriate object segmentation.
However, there is no paper in which an ablation study was conducted to investigate how each loss function influences object segmentation in multi-object representation learning.

To elucidate the mechanism of object segmentation in multi-object representation learning, we conducted an ablation study on MONet \cite{MONet}, which is a typical method.
MONet represents multiple objects using pairs that consist of an attention mask and the latent vector corresponding to the attention mask.
The attention masks are generated from the input image, are mutually exclusive and cover the entire input image region.
Each latent vector is encoded from the input image and attention mask.
Then, the component image and attention mask are decoded from each latent vector.
The loss function of MONet consists of 1) a loss for the reconstruction of the input image, 2) the VAE regularization loss of the latent vector, and 3) the reconstruction loss of the attention mask to explicitly encode shape information.
The first loss contains the sum of reconstruction losses between the input image and the decoded component image.
Additionally, each reconstruction loss is weighted by the attention mask pixel by pixel.
Thus, for each pixel, the smaller the reconstruction loss, the larger the attention mask.
We conducted an ablation study on these three loss functions to investigate the effect on segmentation performance.

Our contributions can be summarized as follows:
{\setlength{\leftmargini}{10pt}\begin{itemize}\vspace{-4pt}
  \item We evaluated the change in object segmentation performance when each loss is removed or replaced.
        Our results showed that the VAE regularization loss does not significantly affect segmentation performance, and the other losses did affect it.
  \item Based on this result, we hypothesize that it is important to maximize the attention mask of the image region best represented by a single latent vector corresponding to the attention mask.
        To confirm this hypothesis, we designed a new loss function that has the same mechanism as the hypothesis.
        Our results showed that the new loss function did not degrade segmentation performance compared with the original loss function, thus confirming the hypothesis.
\end{itemize}}

\section{\uppercase{Background:MONet}}\label{background}

In this section we explain the loss function of MONet \cite{MONet} in detail.
Then we briefly explain the inference process and the definition of variables in MONet.
MONet represents an input image $\mathbf x \in \mathbb R^{H \times W \times C}$ using $K$ pairs of an attention mask $\mathbf m_k \in \mathbb R^{H \times W \times 1}$, where $k \in 1,...,K$ and the corresponding latent vector $\mathbf z_k \in \mathbb R^{D}$.
The architecture of MONet consists of an attention module and component VAE.

The attention module generates the attention masks $\mathbf m_1,...,\mathbf m_K$ from the input image $\mathbf x$.
The purpose of the attention masks is to decompose the input image $\mathbf x$ into multiple elements.
The attention masks are constrained to have a range of values $[0,1]$ and to satisfy $\sum_{k = 1}^K\mathbf m_k = \mathbf 1$ for each pixel.
Under these constraints, the attention masks are mutually exclusive and cover the entire input image region.

\begin{figure}\centering\includegraphics[clip,width=.4\textwidth]{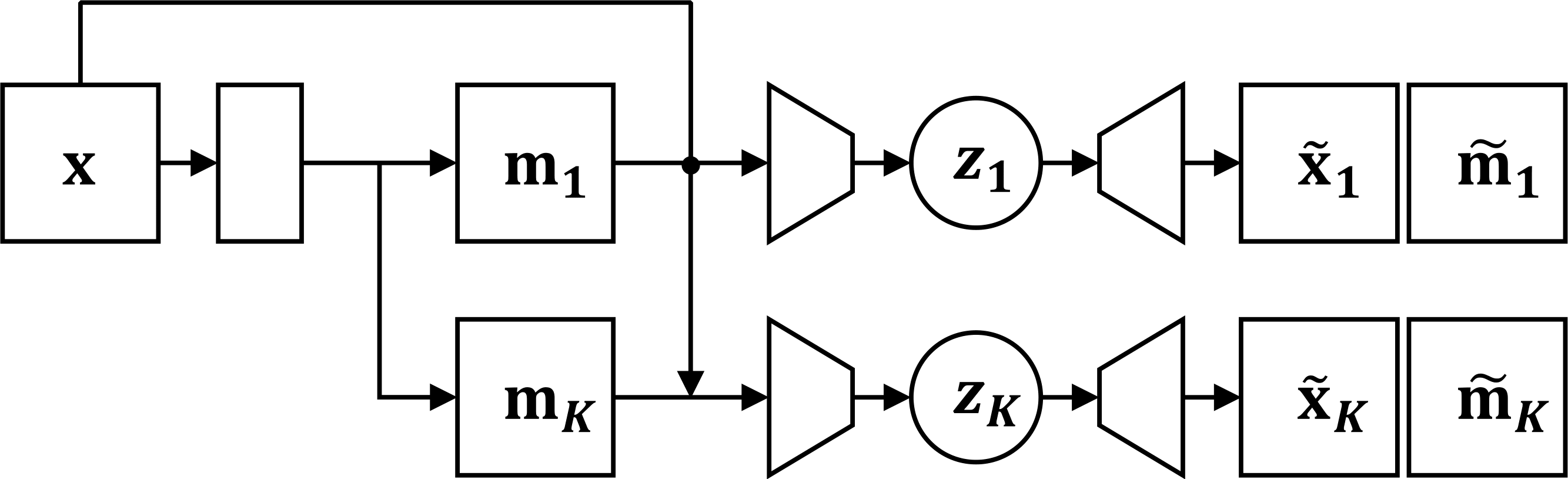}\caption{
  Schematic of the inference of MONet \cite{MONet}.
  Given an input image $\mathbf x$, attention masks $\mathbf m_1,...,\mathbf m_K$ are generated.
  Then, latent vectors $\mathbf z_1,...,\mathbf z_K$ are inferred.
  Finally, component images $\tilde{\mathbf x}_1,...,\tilde{\mathbf x}_K$ and reconstructed attention masks $\tilde{\mathbf m}_1,...,\tilde{\mathbf m}_K$ are generated.
}\label{fig:MONet}\end{figure}

The component VAE consists of an encoder and decoder.
The latent vectors and decoded images of the component VAE are based on a probabilistic formulation, as is VAE \cite{VAE}.
The encoder infers the mean and standard deviation $\boldsymbol \mu_k, \boldsymbol \sigma_{z,k}$ of the posterior distribution of the $k$th latent vector $q(\mathbf z_k|\mathbf x,\mathbf m_k)$ from the input image $\mathbf x$ and $k$th attention mask $\mathbf m_k$.
In the encoding process, the $k$th attention mask $\mathbf m_k$ works as an indicator of which regions of the input image $\mathbf x$ should be encoded.
The $k$th latent vector $\mathbf z_k$ is sampled from the posterior distribution as follows:
\begin{equation}\label{eq:latent_resample}\mathbf z_k = \boldsymbol \mu_k + \boldsymbol \epsilon \boldsymbol \sigma_{z,k},\end{equation}
where $\boldsymbol \epsilon \sim Normal(0,1)$.
The decoder generates the $k$th component image $\tilde{\mathbf x}_k$ and $k$th reconstructed attention mask $\tilde{\mathbf m}_k$ from the $k$th latent vector $\mathbf z_k$.
The $k$th component image $\tilde{\mathbf x}_k$ is intended to reconstruct the region of the input image $\mathbf x$ indicated by the $k$th attention mask $\mathbf m_k$.
The $k$th reconstructed attention mask $\tilde{\mathbf m}_k$ is intended to reconstruct the $k$th attention mask $\mathbf m_k$.
The reconstructed attention masks are also designed to satisfy $\sum_{k = 1}^K\tilde{\mathbf m}_k = \mathbf 1$.
We show the schematic of the inference of MONet in Figure~\ref{fig:MONet}.

The loss function of MONet consists of a loss for the reconstruction of the input image $\mathcal{L}_{nll}$, the VAE regularization loss $\mathcal{L}_{l}$, and the reconstruction loss for the attention mask $\mathcal{L}_{m}$.
$\mathcal{L}_{nll}$ is the negative log likelihood (NLL) designed based on a probabilistic formulation.
In this formulation, the $k$th attention mask $\mathbf m_k$ is considered as the probability that a certain image region belongs to the $k$th component.
The $k$th component image $\tilde{\mathbf x}_k$ is considered as the mean of the $k$th posterior distribution $p(\mathbf x|\mathbf z_k)$ that is considered as a pixel-wise independent Gaussian distribution.
The $k$th posterior distribution $p(\mathbf x|\mathbf z_k)$ is weighted by the $k$th attention mask $\mathbf m_k$, such that it is unconstrained outside the image region that $\mathbf m_k$ indicates.
Then the NLL loss $\mathcal{L}_{nll}$ is formulated to maximize the probability that the input image $\mathbf x$ is sampled from the posterior distribution generated by the decoder as follows:
\begin{equation}\begin{split}
  \mathcal{L}_{nll} &= -\log\sum_{k = 1}^K\mathbf m_kp(\mathbf x\mid\mathbf z_k) \\
                    &= -\log\sum_{k = 1}^K\mathbf m_k\frac{1}{\sigma_{x, k}\sqrt{2\pi}}\exp\left(-\frac{(\mathbf x-\tilde{\mathbf x}_k)^2}{2\sigma_{x, k}^2}\right) \\
                    &= -\log\frac{1}{\sqrt{2\pi}}\sum_{k = 1}^K\exp\left(\log\frac{\mathbf m_k}{\sigma_{x, k}}-\frac{(\mathbf x-\tilde{\mathbf x}_k)^2}{2\sigma_{x, k}^2}\right) \\
                    &= -\sum_{i = 1}^I\log\frac{1}{\sqrt{2\pi}}\sum_{k = 1}^K\exp\left(\log\frac{\mathbf m_k}{\sigma_{x, k}}-\frac{(\mathbf x-\tilde{\mathbf x}_k)^2}{2\sigma_{x, k}^2}\right),
\end{split}\label{NLL_formulation}\end{equation}
where $\sigma_{x,k}$ is the standard deviation of the $k$th posterior distribution $p(\mathbf x|\mathbf z_k)$ and is a scalar constant.
The last line of Equation~\ref{NLL_formulation} means that the NLL is summed over all $I$ pixels of the image, where $I=H \times W \times C$.
In the second line of Equation~\ref{NLL_formulation}, when the $k$th reconstruction loss $(\mathbf x-\tilde{\mathbf x}_k)^2$ decreases, $\mathcal{L}_{nll}$ also decreases as a result of an increase in the exponential term $\exp(-(\mathbf x-\tilde{\mathbf x}_k)^2/{2\sigma_{x, k}^2})$.
Thus, the NLL loss $\mathcal{L}_{nll}$ can be regarded as the pixel-wise weighted sum of the reconstruction losses using the attention masks.

We note that the NLL loss $\mathcal{L}_{nll}$ has a function that makes an integrated reconstructed image $\tilde{\mathbf x}$ closer to the input image $\mathbf x$, where $\tilde{\mathbf x}$ is defined as follows:
\begin{equation}\tilde{\mathbf x} = \sum_{k=1}^{K} \mathbf m_k \tilde{\mathbf x}_k.\end{equation}
In the second line of Equation~\ref{NLL_formulation}, both the attention mask $\mathbf m_k$ and exponential term $\exp(-(\mathbf x-\tilde{\mathbf x}_k)^2/{2\sigma_{x, k}^2})$ are non-negative, and the maximum value is $1$.
If the exponential term is the maximum value of $1$ for pixels for which the attention mask $\mathbf m_k$ is non-zero, the NLL loss $\mathcal{L}_{nll}$ is minimized.
When the exponential term is the maximum value of $1$, the reconstruction loss between the input image $\mathbf x$ and the $k$th component image $\tilde{\mathbf x}_k$ is $0$.
Thus, for pixels for which the attention mask $\mathbf m_k$ is non-zero, the values of the input image $\mathbf x$ and integrated reconstruction (IR) image $\tilde{\mathbf x}$ match.
Additionally, $\sum_{k = 1}^K\mathbf m_k = \mathbf 1$.
Therefore, when the NLL loss $\mathcal{L}_{nll}$ is minimized, $\mathbf x$ and $\tilde{\mathbf x}$ are equal.

The VAE regularization loss $\mathcal{L}_{l}$ is the Kullback--Leibler (KL) divergence \cite{KL} of the posterior distribution of the latent vectors.
$\mathcal{L}_{l}$ is formulated as follows:
\begin{equation}\mathcal{L}_{l} = D_{KL}(\prod_{k=1}^K q(\mathbf z_k|\mathbf x,\mathbf m_k)\parallel p(\mathbf z)),\end{equation}
where $p(\mathbf z)$ is a prior.

The reconstruction loss of the attention mask $\mathcal{L}_{m}$ is the KL divergence between the $k$th attention mask $\mathbf m_k$ and the $k$th reconstructed attention mask $\tilde{\mathbf m}_k$.
We call $\mathcal{L}_{m}$ the mask reconstruction loss.
$\mathcal{L}_{m}$ is formulated as follows:
\begin{equation}\mathcal{L}_{m} = \sum_{k=1}^K D_{KL}(\mathbf m_k\parallel \tilde{\mathbf m}_k)\end{equation}

Then, the entire system of MONet is trained end-to-end using loss $\mathcal{L}$ given by
\begin{equation}\mathcal{L} = \mathcal{L}_{nll}+\beta\mathcal{L}_{l}+\gamma\mathcal{L}_{m},\end{equation}
where $\beta$ and $\gamma$ are hyperparameters used for weighting losses.

\section{\uppercase{Experiments}}
We conducted an ablation study on the three loss functions of MONet to investigate the effect on segmentation performance.
However, wo could not remove the NLL loss $\mathcal{L}_{nll}$ because it is essential for reconstructing the input image $\mathbf x$.
If MONet does not learn to reconstruct the input image $\mathbf x$, then the attention module does not learn appropriate object segmentation.
Thus, we separated the experiment on the NLL loss $\mathcal{L}_{nll}$ from the experiment on the other two losses $\mathcal{L}_{l}$ and $\mathcal{L}_{m}$.
First, we describe the ablation study on the two loss $\mathcal{L}_{l}$ and $\mathcal{L}_{m}$ in Section~\ref{exp_beta_gamma}, and then describe the experiment on the NLL loss $\mathcal{L}_{nll}$ in Section~\ref{exp_mse_loss} and \ref{exp_new_loss}.
Before we explain each experiment, we describe the settings common to all experiments.

\paragraph{Datasets}
We conducted the experiments on Multi-dSprites and ObjectsRoom \cite{ARI_Implement}.
Both datasets were used in the MONet paper \cite{MONet}.
Each dataset contains 1,000,000 images and we withheld 1,000 images to calculate the segmentation metrics.
According to the original settings in the MONet paper, we set the size of the image to $64\times64$ for both datasets and standardized the pixel values of the image from $[0,255]$ to $[-1,1]$.

\paragraph{Hyperparameters}
We describe the values of $\beta$ and $\gamma$ in the section for each experiment.
We set the standard deviation of the $k$th posterior distribution $\sigma_{x,k}$ to $0.11$ when $k=1$; otherwise, it was $0.09$, which was used for the original MONet \cite{MONet}.
We followed the other hyperparameters used for the original MONet.

\paragraph{Training setup}
The original MONet was trained for $1,000,000$ iterations.
However, the training time was too long to ensure experimental results with a sufficient number of random seeds.
Furthermore, in most cases, the number of iterations at which loss functions and segmentation metrics converge is less than $1,000,000$.
We terminated training when convergence was achieved by monitoring the mean squared error (MSE) for the integrated reconstructed image $\tilde{\mathbf x}$ as follows:
\begin{equation}\label{eq:mse_loss}\mathcal{L}_{mse} = \frac{1}{I}\sum_{i = 1}^I (\mathbf x - \tilde{\mathbf x})^2.\end{equation}
When the NLL loss $\mathcal{L}_{nll}$ is at its minimum, $\mathcal{L}_{mse}$ is also at its minimum value of $0$.
We monitored $\mathcal{L}_{mse}^t$, that is, the MSE loss at iteration step $t$, every $100,000$ steps and terminated training when $\mathcal{L}_{mse}^t<l$ and $\mathcal{L}_{mse}^t/\mathcal{L}_{mse}^{t-100,000}>0.99$.
We set threshold $l$ to $0.005$ when the dataset was ObjectsRoom and to $0.001$ when the dataset was Multi-dSprites.
In each experiment, we confirmed, if necessary, that the influence of the termination was sufficiently small.
For the remaining training setup, such as the optimizer and learning rate, we followed the original MONet.

\begin{figure}[h]\centering
  \begin{minipage}[t]{0.32\linewidth}\centering\includegraphics[width=1.0\linewidth]{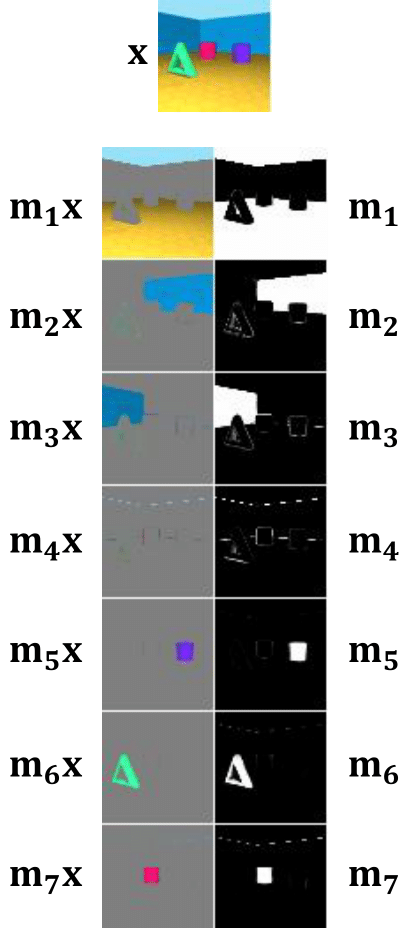}\vspace{0pt}\subcaption{ARI$=0.7$}\end{minipage}
  \begin{minipage}[t]{0.32\linewidth}\centering\includegraphics[width=1.0\linewidth]{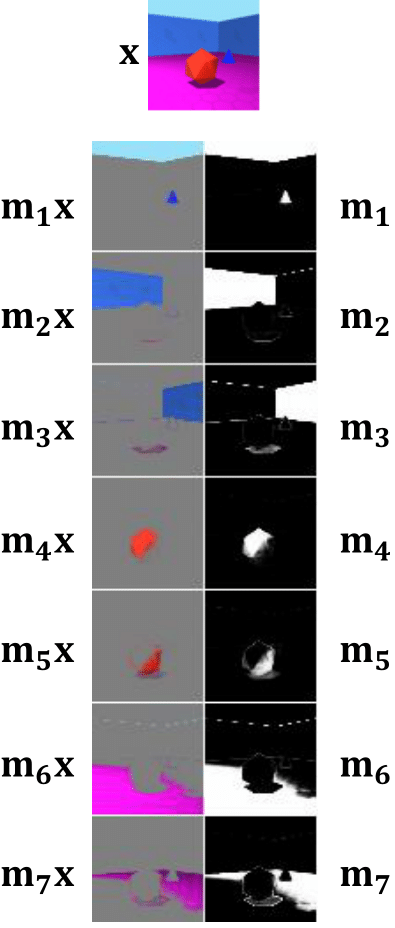}\vspace{0pt}\subcaption{ARI$=0.5$}\end{minipage}
  \begin{minipage}[t]{0.32\linewidth}\centering\includegraphics[width=1.0\linewidth]{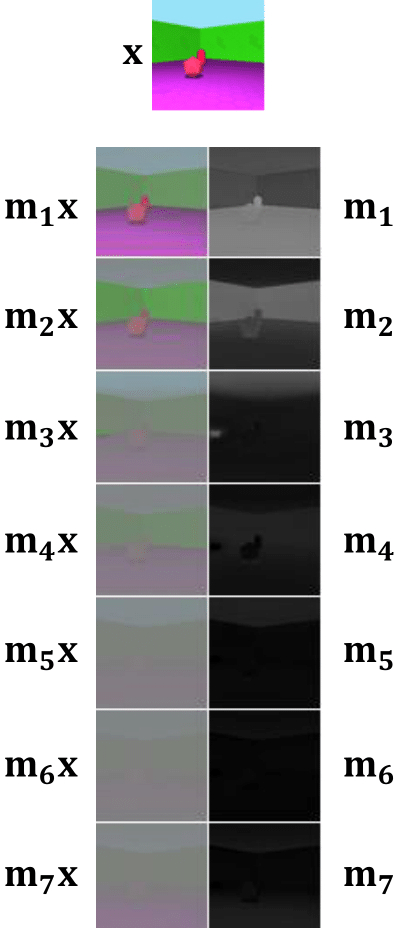}\vspace{0pt}\subcaption{ARI$=0.3$}\end{minipage}
  \vspace{0pt}\caption{
    Example of attention masks at (a) ARI$=0.7$, (b) ARI$=0.5$, and (c) ARI$=0.3$ for ObjectsRoom.
    In (a), the foreground objects are appropriately segmented.
    In (b), a red object split in two.
    In (c), attention masks are not binarized.
}\label{fig:ARI_masks}\end{figure}

\paragraph{Evaluation}
We provided $100$ random seeds from $1$ to $100$ for each experimental condition for the quantitative evaluation.
We quantified segmentation performance using the adjusted Rand index (ARI) \cite{ARI}; the larger the values the better.
When the attention mask matches the ground-truth segmentation masks, the ARI takes the maximum value of 1.
In the ObjectsRoom dataset, there were background components, such as the sky, wall, and floor.
However, it was not obvious whether there was a single appropriate background segmentation, such as whether the wall and floor should be separated.
Therefore, we removed the background masks from the ground-truth masks and then evaluated whether the foreground objects were appropriately segmented.
We used the metrics at the end of training for evaluation.
Example of attention masks at ARI is 0.7, 0.5, and 0.3 for ObjectsRoom is shown in Figure~\ref{fig:ARI_masks}.

\paragraph{Computation}
We conductd a single training process using a single NVIDIA A100 GPU.
A single training process took 20 hours, on average.
We implemented the MONet model using PyTorch \cite{PyTorch}.

\subsection{Ablation study on $\beta$ and $\gamma$ of the loss function}\label{exp_beta_gamma}

We investigated the influence on segmentation performance when we removed the VAE regularization loss $\mathcal{L}_{l}$ and mask reconstruction loss $\mathcal{L}_{m}$.
Segmentation performance degrades if the loss function plays an important role.
We removed loss function by setting the weight for each loss function $\beta$ and $\gamma$ to $0$.
Then, we set up four experimental conditions $[\mathbf{11,01,10,00}]$ as follows:

\centerline{$\begin{matrix*}[l]
  \mathbf{11} & \beta=0.5 & \gamma=0.25 \\
  \mathbf{01} & \beta=0   & \gamma=0.25 \\
  \mathbf{10} & \beta=0.5 & \gamma=0    \\
  \mathbf{00} & \beta=0   & \gamma=0.    \\
\end{matrix*}$}

The condition $\mathbf{11}$ indicates the original MONet settings.
When $\beta$ was $0$, we set $\boldsymbol \epsilon$ in Equation~\ref{eq:latent_resample} to $\mathbf 0$ to avoid the VAE's resampling \cite{VAE}.

\begin{figure}[h]\centering
  \begin{minipage}[t]{0.49\linewidth}\centering\includegraphics[width=1.0\linewidth]{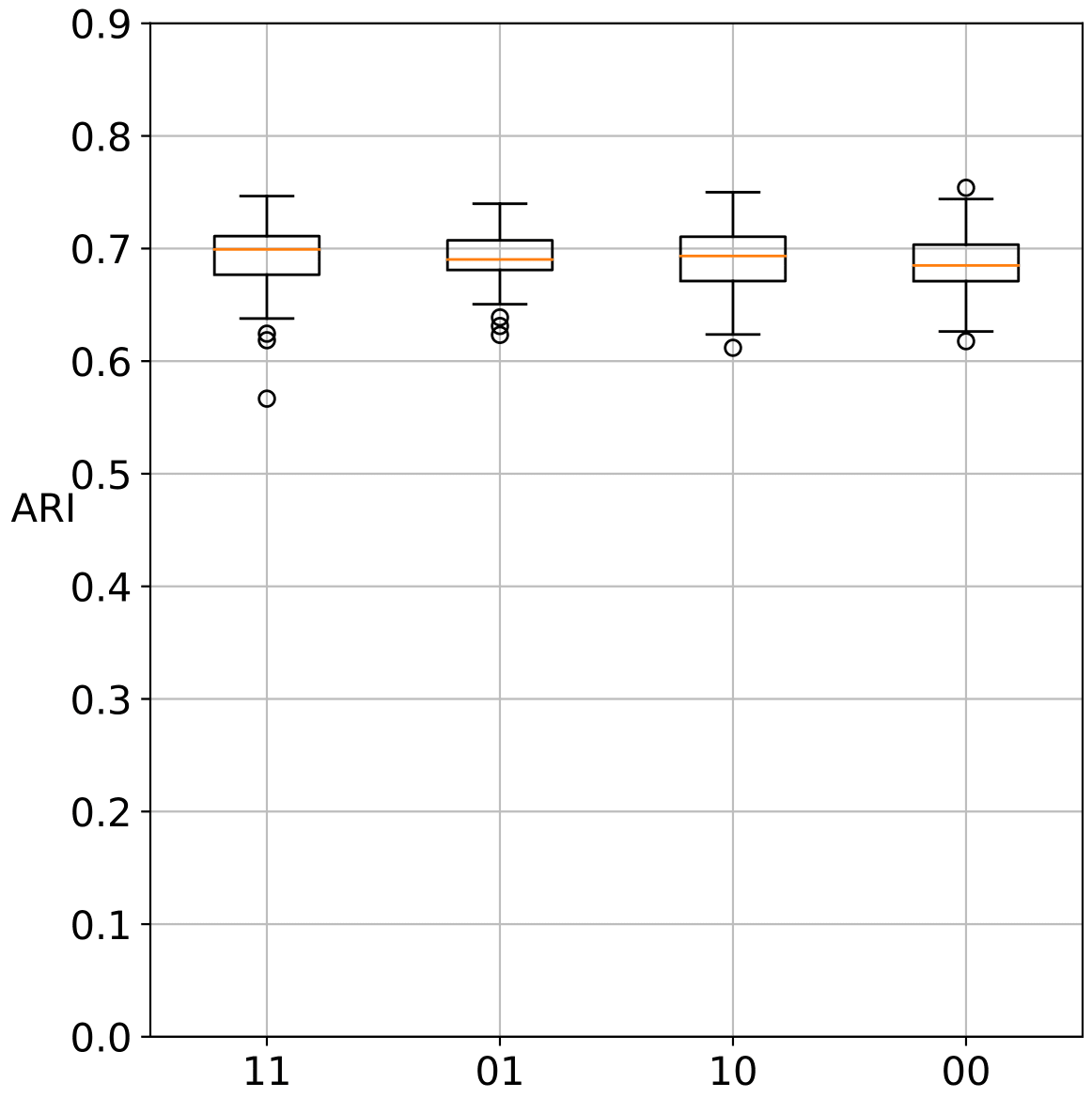}\vspace{0pt}\subcaption{Multi-dSprites}\label{fig:exp1_compare_multi_dsprites}\end{minipage}
  \begin{minipage}[t]{0.49\linewidth}\centering\includegraphics[width=1.0\linewidth]{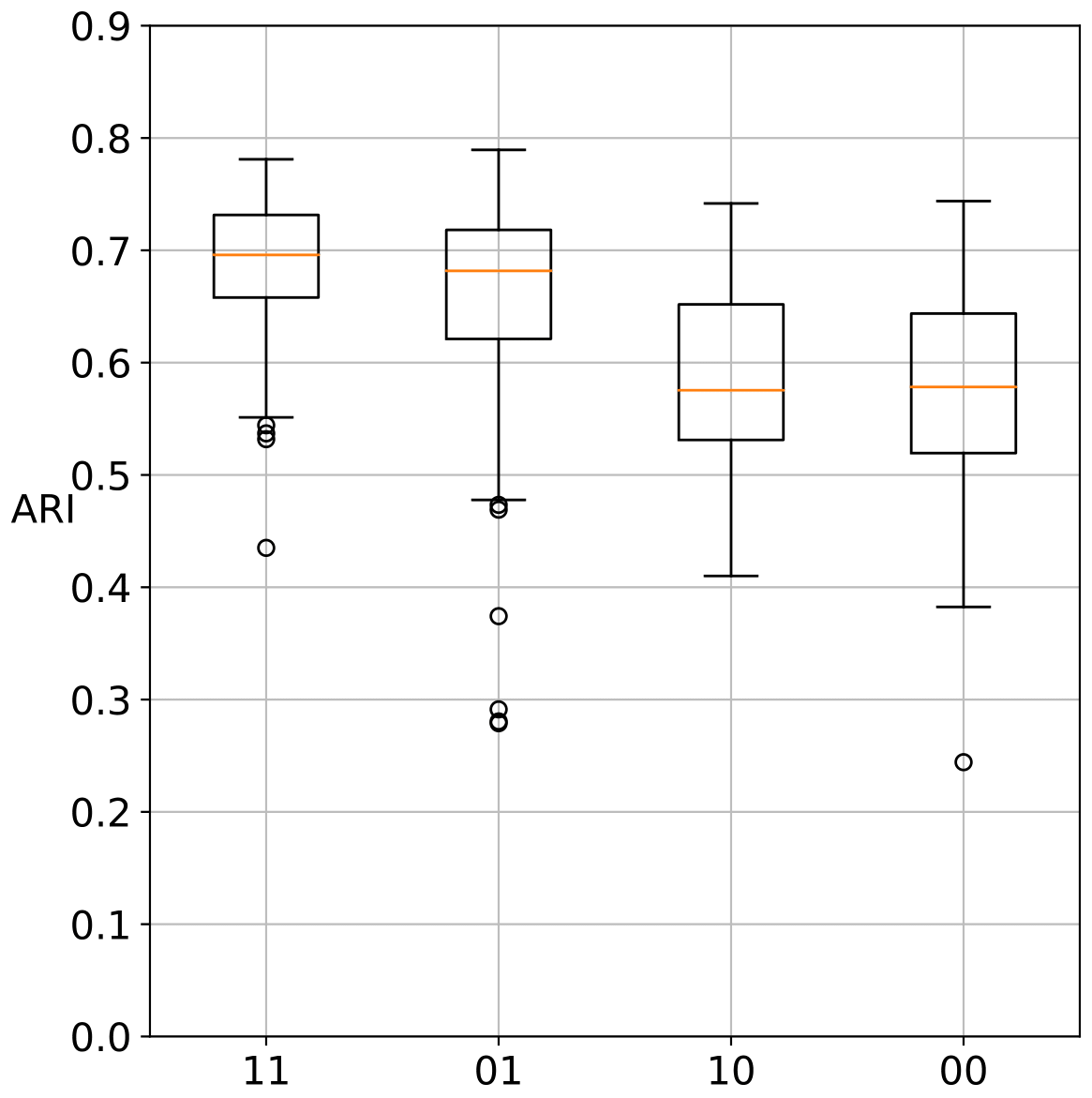}\vspace{0pt}\subcaption{ObjectsRoom}\label{fig:exp1_compare_objects_room}\end{minipage}
  \vspace{0pt}\caption{
    Segmentation performance under each condition $[\mathbf{11,01,10,00}]$ in the ablation study in Section~\ref{exp_beta_gamma}.
    Blank circles represent outliers.
    Based on the comparison of conditions $\mathbf{11}$ and $\mathbf{01}$, or $\mathbf{10}$ and $\mathbf{00}$, the difference in $\beta$ did not change the average value of the segmentation metrics.
    For ObjectsRoom, when $\gamma$ was $0$, segmentation performance degraded.
}\label{fig:exp1_compare}\end{figure}

Segmentation performance under each condition is shown in Figure~\ref{fig:exp1_compare}.
For Multi-dSprites, the average segmentation performance was almost the same under all conditions.
For ObjectsRoom, when $\gamma=0$, that is, when we removed the mask reconstruction loss $\mathcal{L}_{m}$, the average segmentation performance degraded.
Additionally, when $\beta$ changed, the average segmentation performance did not change substantially.

We also evaluated whether the presence of each loss affected segmentation performance using a statistical test.
We performed Friedman's test \cite{FriedmanTest}.
We set the limit value of the $p$-value to $0.05$.
For Multi-dSprites, the medians of the segmentation metircs in condition $[\mathbf{11,01,10,00}]$ were $[\mathbf{0.70,0.69,0.69,0.68}]$.
And then, the $p$-value about $\beta$ was $0.32$ and the $p$-value about $\gamma$ was $0.048$.
For ObjectsRoom, the medians of the segmentation metircs in condition $[\mathbf{11,01,10,00}]$ were $[\mathbf{0.70,0.68,0.58,0.58}]$.
And then, the $p$-value for $\beta$ was $0.16$ and the $p$-value for $\gamma$ was $\num{1.7E-22}$.
Thus, for both datasets, the change of $\beta$ was not significantly different and setting $\gamma$ to $0$ significantly degraded segmentation performance.

\begin{table*}[h]\caption{
  Difference between the segmentation metrics at the moment the training is terminated according to our criteria and those after full training.
  The ratio is calculated by dividing the difference by the median of the segmentation metric.
  These values are averages over five random seeds.
  Compared with the size of the segmentation metrics in Figure~\ref{fig:exp1_compare}, these differences are sufficiently small.
}\label{tab:ari_differences}\centering
  \begin{minipage}[c]{1\linewidth}\centering\subcaption{Multi-dSprites}\renewcommand{\arraystretch}{1.2}\begin{tabular}{|l||c|c|c|c|}
    \hline
    condition & $\mathbf{11}$ & $\mathbf{10}$ & $\mathbf{01}$ & $\mathbf{00}$ \\
    \hline\hline
    difference &$\num{-4.5E-4}$ & $\num{-3.5E-4}$ & $\num{2.3E-3}$ & $\num{2.6E-3}$ \\
    \hline
    ratio(\%)  &$-0.064$ & $-0.050$ & $0.33$ & $0.38$ \\
    \hline
  \end{tabular}\end{minipage}\vspace{10pt}
  \begin{minipage}[c]{1\linewidth}\centering\subcaption{ObjectsRoom}\renewcommand{\arraystretch}{1.2}\begin{tabular}{|l||c|c|c|c|}
    \hline
    condition & $\mathbf{11}$ & $\mathbf{10}$ & $\mathbf{01}$ & $\mathbf{00}$ \\
    \hline\hline
    difference & $\num{-2.2E-4}$ & $\num{4.9E-3}$ & $\num{-2.8E-3}$ & $\num{2.8E-3}$ \\
    \hline
    ratio(\%)  & $-0.032$ & $0.86$ & $-0.40$ & $0.48$ \\
    \hline
  \end{tabular}\end{minipage}\hfill
\end{table*}

Finally, we confirmed that the effect on the experimental results of terminating the process in the middle of training was sufficiently small.
We performed training for the maximum training iterations is performed using five random seeds for each experimental condition.
Then, we calculated the difference between the value of the segmentation metrics at the time the process was terminated by our criteria and the final value of the metrics.
The differences are shown in Table~\ref{tab:ari_differences}.
Compared with the size of the segmentation metrics, the differences were sufficiently small.

These results indicate that the VAE regularization loss $\mathcal{L}_{l}$ did not significantly affect segmentation performance, whereas the mask reconstruction loss $\mathcal{L}_{m}$ did.

\subsection{Investigating the role of the NLL loss in object segmentation}\label{exp_mse_loss}
In Section~\ref{background}, we noted that the NLL loss $\mathcal{L}_{nll}$ had a function to that makes the integrated reconstructed image $\tilde{\mathbf x}$ closer to the input image $\mathbf x$.
However, it is not clear whether the NLL loss has the same function as a mere reconstruction loss between the input image $\mathbf x$ and the integrated reconstructed $\tilde{\mathbf x}$ image for appropriate object segmentation.
In this section, we replace the NLL loss $\mathcal{L}_{nll}$ with another reconstruction loss $\mathcal{L}_{ir}$.
In the formulation of $\mathcal{L}_{ir}$, we consider that the integrated reconstructed image $\tilde{\mathbf x}$ is the mean of the Gaussian posterior distribution $r(\mathbf x\mid\mathbf z_k,\mathbf m_k)$.
The formulation of $\mathcal{L}_{ir}$ is as follows:
\begin{equation}\begin{split}
  \mathcal{L}_{ir}  &= -\log r(\mathbf x\mid\mathbf z_k,\mathbf m_k) \\
                    &= -\log\frac{1}{\sigma_{x}\sqrt{2\pi}}\exp\left(-\frac{(\mathbf x-\tilde{\mathbf x})^2}{2\sigma_{x}^2}\right) \\
                    &= -\log\frac{1}{\sigma_{x}\sqrt{2\pi}}+\sum_{i = 1}^I\frac{(\mathbf x-\tilde{\mathbf x})^2}{2\sigma_{x}^2}.
\end{split}\label{IR_formulation}\end{equation}
Therefore, $\mathcal{L}_{ir}$ can be regarded as a linear transformation of the MSE loss $\mathcal{L}_{mse}$ in Equation~\ref{eq:mse_loss}.
We call $\mathcal{L}_{ir}$ the IR loss.
The function of the IR loss $\mathcal{L}_{ir}$ is to reconstruct the input image, but it does not include the mask weighting of the $k$th reconstruction loss $(\mathbf x-\tilde{\mathbf x}_k)^2$ like the NLL loss $\mathcal{L}_{nll}$.

In the experiment, we set up two experimental conditions: NLL+M and IR+M.
The condition NLL+M means that we used the NLL loss $\mathcal{L}_{nll}$ and mask reconstruction loss $\mathcal{L}_{m}$, which is the same as condition $\mathbf{01}$ in the previous section.
Because we showed in the previous section that the mask reconstruction loss is involved in object segmentation, we applied the mask reconstruction loss.
The condition IR+M means that we used the IR loss $\mathcal{L}_{ir}$ and mask reconstruction loss $\mathcal{L}_{m}$.
We set $\beta$ to $0$ and $\gamma$ to $0.25$.
We set the standard deviation of the posterior distribution $\sigma_{x}$ to $0.09$.

\begin{figure}[h]\centering
  \begin{minipage}[t]{0.49\linewidth}\centering\includegraphics[width=1.0\linewidth]{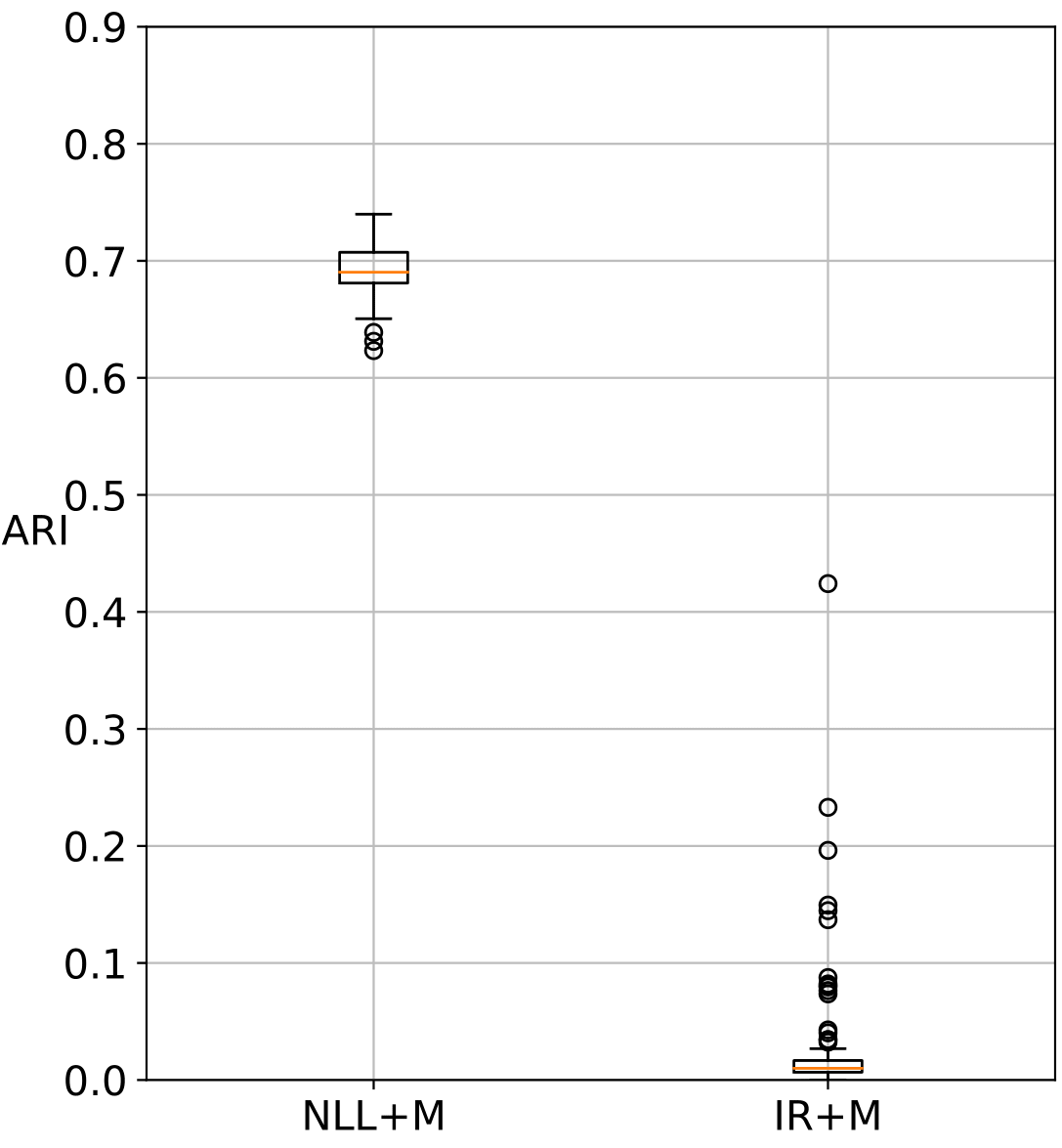}\vspace{0pt}\subcaption{Multi-dSprites}\label{fig:exp2_compare_multi_dsprites}\end{minipage}
  \begin{minipage}[t]{0.49\linewidth}\centering\includegraphics[width=1.0\linewidth]{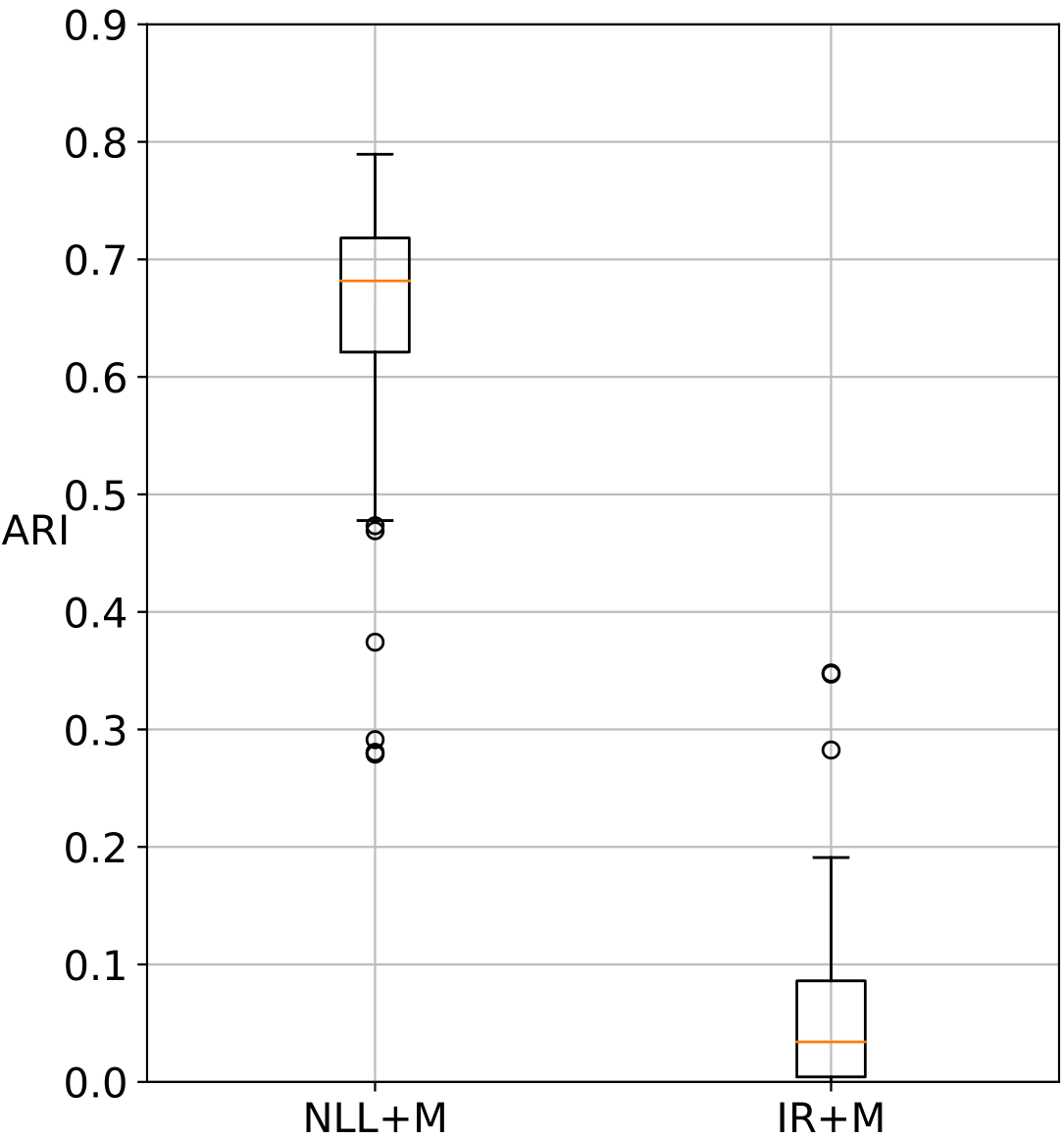}\vspace{0pt}\subcaption{ObjectsRoom}\label{fig:exp2_compare_objects_room}\end{minipage}
  \vspace{0pt}\caption{
    Segmentation performance for each condition NLL+M and IR+M in Section~\ref{exp_mse_loss}.
    Blank circles represent outliers.
    For both datasets, the replacement of the NLL loss with the IR loss degraded segmentation performance.
}\label{fig:exp2_compare}\end{figure}

Segmentation performance under each condition is shown in Figure~\ref{fig:exp2_compare}.
We also evaluated whether there is a significant difference between segmentation performance under the two conditions.
We performed Wilcoxon signed-rank test \cite{WilcoxonTest}.
We set the limit value of the $p$-value to $0.05$.
For Multi-dSprites, the medians of the segmentation metircs in condition NLL+M and IR+M were $0.69$ and $0.010$.
And then the $p$-value was $0$.
For ObjectsRoom, the medians of the segmentation metircs in condition NLL+M and IR+M were $0.68$ and $0.034$.
And then the $p$-value was $0$.
Thus, for both datasets, the replacement of the NLL loss with the IR loss significantly degraded segmentation performance.
From this result, we inferred that the NLL loss is critically important for appropriate object segmentation.

\subsection{Replacement of the new loss function with the same mechanism as the NLL loss}\label{exp_new_loss}
Based on the experimental results, we hypothesize that it is important to maximize the attention mask of the image region best represented by a single latent vector $\mathbf z_k$ corresponding to the attention mask $\mathbf m_k$.
The NLL loss $\mathcal{L}_{nll}$ can be regarded as a pixel-wise weighted sum of the reconstruction losses by the attention masks.
Because the attention masks are restricted to satisfy $\sum_{k = 1}^K\mathbf m_k = \mathbf 1$, all values of the attention masks cannot be zero.
Then the attention masks have to reflect at least one reconstruction loss from $K$ reconstruction losses $(\mathbf x-\tilde{\mathbf x}_1)^2,...,(\mathbf x-\tilde{\mathbf x}_K)^2$ in the NLL loss $\mathcal{L}_{nll}$.
In this scenario, for each pixel, the loss function is minimized if the attention mask has a maximum value of $1$ for only $k$ that has the smallest reconstruction loss $(\mathbf x-\tilde{\mathbf x}_k)^2$.
Thus, we hypothesize that weighting the reconstruction loss with the attention mask $\mathbf m_k$ naturally causes the attention mask $\mathbf m_k$ to become binarized and focus on the region that can be represented by a single latent vector $\mathbf z_k$, that is, a single object.

To confirm this hypothesis, we designed a new loss function $\mathcal{L}_{mw}$ that has the same mechanism as the NLL loss $\mathcal{L}_{nll}$ based on our hypothesis.
The formulation of $\mathcal{L}_{mw}$ is as follows:
\begin{equation}\label{eq:mw_loss}\mathcal{L}_{mw} = \frac{1}{I}\sum_{i = 1}^I \sum_{k=1}^K \mathbf m_k(\mathbf x - \tilde{\mathbf x}_k)^2\end{equation}
We call $\mathcal{L}_{mw}$ the mask-weighted (MW) loss.
We also formulated a new mask reconstruction loss $\mathcal{L}_{m}^{new}$ in conjunction with the formulation of the MW loss $\mathcal{L}_{mw}$ as follows:
\begin{equation}\label{eq:new_mask_loss}\mathcal{L}_{m}^{new} = \frac{1}{I}\sum_{i = 1}^I \sum_{k=1}^K (\mathbf m_k - \tilde{\mathbf m}_k)^2.\end{equation}

We set up three experimental conditions: MSE+M, MW+M, and NLL+M.
The condition MSE+M means that we used the MSE loss $\mathcal{L}_{mse}$ and new mask reconstruction loss $\mathcal{L}_{m}^{new}$.
MSE+M is an ablation condition for MW+M, as well as IR+M for NLL+M in Section~\ref{exp_mse_loss}.
To be consistent with the formulation of the MW loss $\mathcal{L}_{mw}$, we used the MSE loss $\mathcal{L}_{mse}$ instead of the IR loss $\mathcal{L}_{ir}$.
The condition MW+M means that we used the MW loss $\mathcal{L}_{mw}$ and new mask reconstruction loss $\mathcal{L}_{m}^{new}$.
For these two losses, $\gamma = 0.01$ when the dataset was Multi-dSprites and $\gamma = 0.1$ when the dataset was ObjectsRoom.
These $\gamma$ are those that performed best under condition MW+M in the preliminary experiments.
The condition NLL+M was the same as that in Section~\ref{exp_mse_loss}.

\begin{figure}[h]\centering
  \begin{minipage}[t]{0.49\linewidth}\centering\includegraphics[width=1.0\linewidth]{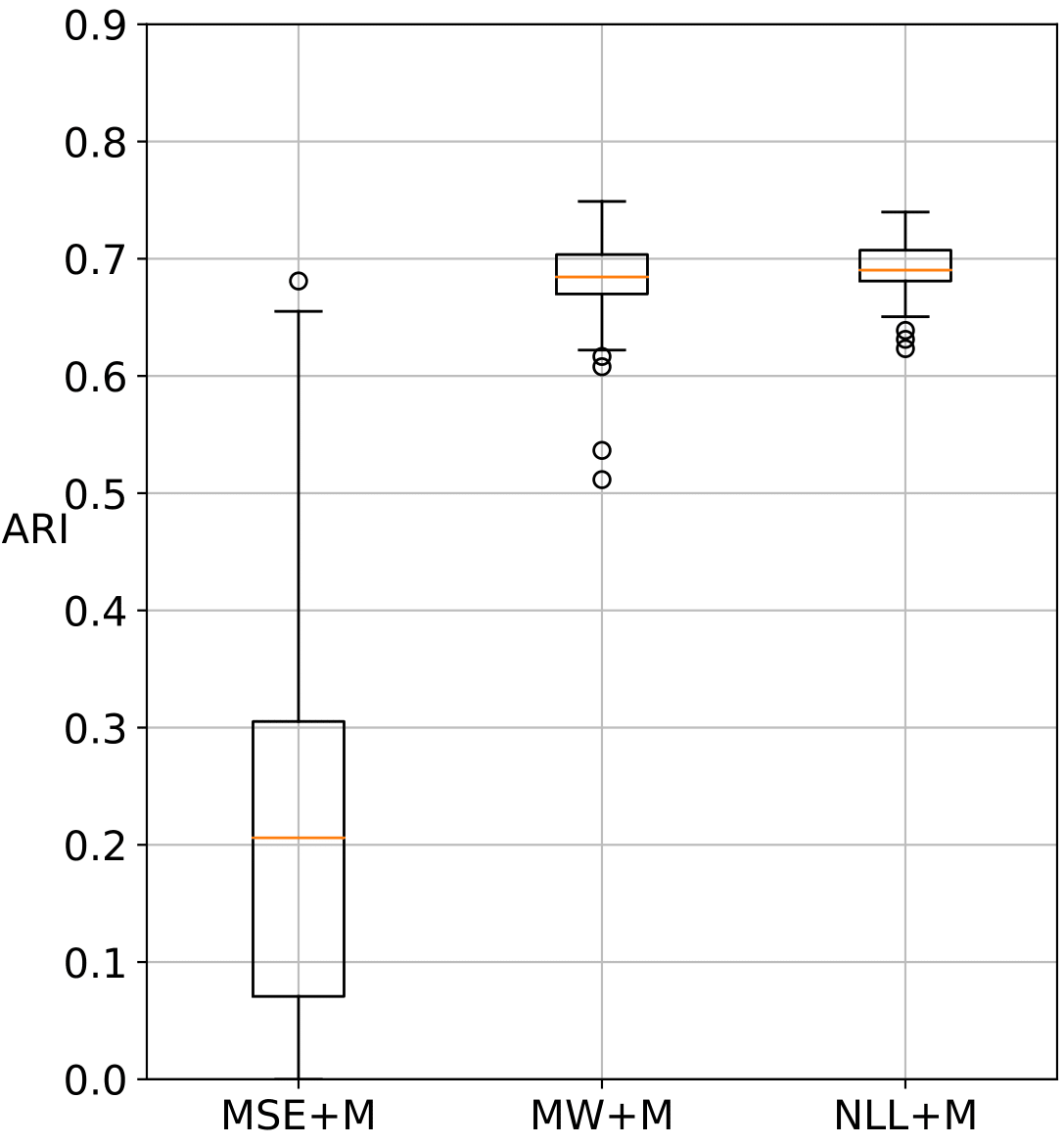}\vspace{0pt}\subcaption{Multi-dSprites}\label{fig:exp3_compare_multi_dsprites}\end{minipage}
  \begin{minipage}[t]{0.49\linewidth}\centering\includegraphics[width=1.0\linewidth]{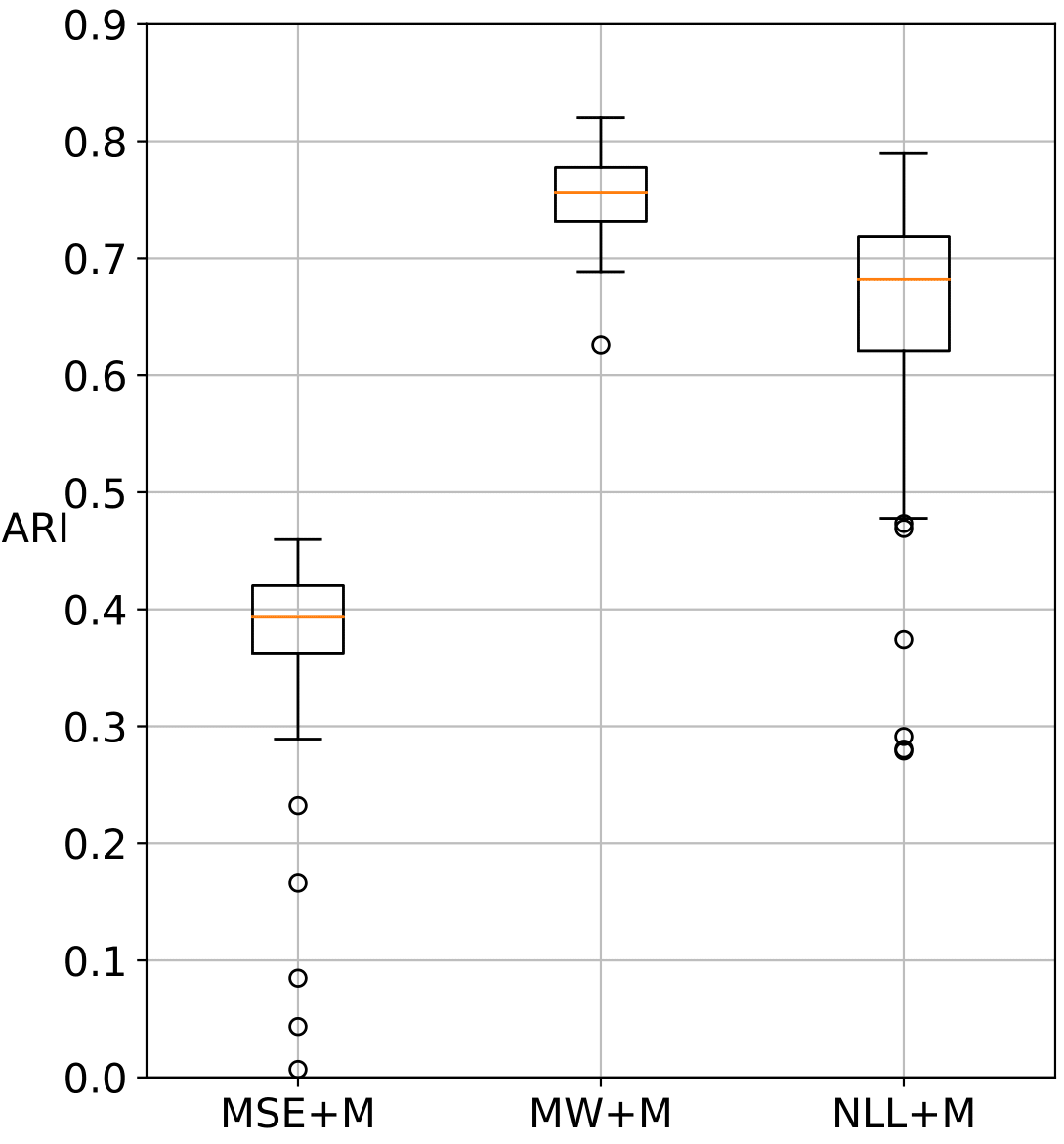}\vspace{0pt}\subcaption{ObjectsRoom}\label{fig:exp3_compare_objects_room}\end{minipage}
  \vspace{0pt}\caption{
    Segmentation performance for each condition MSE+M, MW+M, and NLL+M in Section~\ref{exp_new_loss}.
    Blank circles represent outliers.
    For both datasets, using the MW loss outperformed using the MSE loss.
    Additionally, segmentation performance under the MW+M condition was equal to or better than performance under the NLL+M condition.
}\label{fig:exp3_compare}\end{figure}

Segmentation performance under each condition is shown in Figure~\ref{fig:exp3_compare}.
We also evaluated whether there is a significant difference among segmentation performance under the three conditions.
We performed Wilcoxon signed-rank test \cite{WilcoxonTest} with Holm--Bonferroni method \cite{Bonferroni_Method}.
We conducted three comparisons $[1,2,3]$; 1) [MSE+M, MW+M], 2) [MW+M, NLL+M], and 3) [NLL+M, MSE+M].
We set the limit value of the $p$-value to $0.05/3=0.016$.
For Multi-dSprites, the medians of the segmentation metircs in condition MSE+M, MW+M, and NLL+M were $0.34$, $0.68$, and $0.69$.
And then the $p$-value of each comparison $[1,2,3]$ was $[0,0.019,0]$.
For ObjectsRoom, the medians of the segmentation metircs in condition MSE+M, MW+M, and NLL+M were $0.39$, $0.76$, and $0.68$.
And then the $p$-value of each comparison $[1,2,3]$ was $[0,\num{6.7E-16},0]$.
Thus, for both datasets, using the MW loss $\mathcal{L}_{mw}$ significantly outperformed using the MSE loss $\mathcal{L}_{mse}$.
Additionally, using the MW loss $\mathcal{L}_{mw}$, performance was comparable to or better than performance using the NLL loss $\mathcal{L}_{nll}$.

Based on these results, we considered that the new loss $\mathcal{L}_{mw}$ worked as well as the NLL loss $\mathcal{L}_{nll}$.
Thus, our hypothesis that it is important to maximize the attention mask of the image region best represented by a single latent vector $\mathbf z_k$ corresponding to the attention mask $\mathbf m_k$ is confirmed.

\section{\uppercase{Discussion}}
In this study, we conducted an ablation study on MONet \cite{MONet} to investigate the object segmentation mechanism in multi-object representation learning.
Our results showed that the VAE regularization loss did not significantly affect segmentation performance and other losses did affect it.
Then we hypothesized that it is important to maximize the attention mask of the image region best represented by a single latent vector corresponding to the attention mask.
We confirmed this hypothesis by evaluating a new loss function that had the same mechanism as the hypothesis.

In MONet, the VAE regularization loss $\mathcal{L}_{l}$ is used to disentangle each element of each latent vector $\mathbf z_k$ \cite{Understanding_VAE}.
However, the MONet paper did not consider the relationship between the VAE regularization loss $\mathcal{L}_{l}$ and object segmentation.
For the first time, we conducted an ablation study on the VAE regularization loss $\mathcal{L}_{l}$ in MONet.
Our result contributes to clarify that removing the VAE regularization loss $\mathcal{L}_{l}$ is not critical to segmentation performance.

In the MONet paper, the formulation of weighting the reconstruction loss by the attention mask $\mathbf m_k$ was derived with the goal of ensuring that the latent vector $\mathbf z_k$ is unconstrained to learn to reconstruct the image region where the attention mask $\mathbf m_k$ is $0$.
However, in the early stages of training, all masks are non-zero and the segmentation boundaries are ambiguous, thus the goal is not always valid during training.
Therefore, no mechanism was presented that could consistently hold throughout training and explain the spontaneous binarization of the attention mask $\mathbf m_k$.
We hypothesized that a winner-take-all mechanism among the latent vectors $\mathbf z_1,...\mathbf z_K$ is important, where the attention mask $\mathbf m_k$ of the image region best represented by a single latent vector $\mathbf z_k$ takes the maximum value $1$ and the others takes the minimum value $0$.
For the first time, we explained that the winner-take-all mechanism was derived from minimizing the NLL loss $\mathcal{L}_{nll}$ in MONet.
Additionally, we confirmed this hypothesis by evaluating a new loss function that had the same winner-take-all mechanism as the NLL loss $\mathcal{L}_{nll}$.
Our result contributes in that to clarify the existence of the winner-take-all mechanism in the NLL loss $\mathcal{L}_{nll}$ and its effect in appropriate learning of the attention mask $\mathbf m_k$.

A previous study on GENESIS \cite{GENESIS}, which is an extended model of MONet and whose basic formulation is the same as that of MONet, investigated the mechanism of object segmentation \cite{Analysis_Bottleneck}.
This study investigated the relationship between segmentation performance and the dimension of the latent vector of the component VAE.
This result showed that changing the capacity of a single latent vector to reconstruct the image affect segmentation performance.
However, which loss functions critically affect segmentation performance was still not clear.
Unlike this approach, we investigated the effect of the loss functions of MONet on segmentation performance by combining an ablation study and evaluation with a new loss function.
Our result contributes to clarify that the VAE regularization loss did not significantly affect segmentation performance and the NLL loss critically did affect it.

Other methods \cite{IODINE,SPACE} of multi-object representation learning are also based on the formulation of MONet.
These methods differ from MONet in the way they generate the attention masks.
Therefore, the result of our study is also applicable to other methods of multi-object representation learning.

Our findings are limited to multi-object representation learning methods that include both reconstruction learning with a single latent vector and attention mask learning.
Unsupervised segmentation methods exist that do not involve reconstruction learning, such as methods based on the mutual information between local image regions \cite{Invariant_Clustering}.
The relationship between these mechanisms and ours is still not clear.
A more general understanding of the mechanism of unsupervised segmentation methods is needed in the future.

%

\bibliographystyle{apalike}
{\small
\bibliography{bib/references}}

%

\end{document}